\documentclass{article}

\usepackage{arxiv}

\usepackage[utf8]{inputenc} %
\usepackage[T1]{fontenc}    %
\usepackage{hyperref}       %
\usepackage{url}            %
\usepackage{booktabs}       %
\usepackage{amsfonts}       %
\usepackage{nicefrac}       %
\usepackage{microtype}      %
\usepackage{lipsum}

\usepackage{amsmath}    %
\usepackage{subcaption} %
\usepackage{graphicx}   %
\usepackage{courier}    %

\usepackage{xcolor}     %

\newcommand{\dpoint}{\boldsymbol{x}}
\newcommand{\dpointlat}{\textit{lat}}
\newcommand{\dpointlon}{\textit{lon}}
\newcommand{\dpointtime}{\textit{t}}
\newcommand{\dpointsigma}{\sigma}

\newcommand{\traj}{S}

\newcommand{\gppredset}{\mathcal{T}}

\newcommand{\gpfunc}{f}
\newcommand{\kernelfunc}{k}
\newcommand{\meanfunc}{m}
\newcommand{\locvalue}{v}
\newcommand{\sigmam}{\dpointsigma_m}
\newcommand{\sigmaprior}{\dpointsigma_0}
\newcommand{\fvec}{\boldsymbol{X}}
\newcommand{\lvec}{\boldsymbol{y}}
\newcommand{\normal}{\mathcal{N}}

\newcommand{\model}{m}

\title{Gaussian Process for Trajectories}

\author{
  Kien Nguyen\\
  Computer Science Department\\
  University of Southern California\\
  Los Angeles, CA  90007 \\
  \texttt{kien.nguyen@usc.edu} \\
  \And
  John Krumm\\
  Microsoft Research\\
  Microsoft Corporation\\
  Redmond, WA  98052 \\
  \texttt{jckrumm@microsoft.com} \\
  \And
  Cyrus Shahabi\\
  Computer Science Department\\
  University of Southern California\\
  Los Angeles, CA  90007 \\
  \texttt{shahabi@usc.edu} \\
}

\begin{document}
\maketitle

\begin{abstract}
The Gaussian process is a powerful and flexible technique for interpolating spatiotemporal data, especially with its ability to capture complex trends and uncertainty from the input signal. This chapter describes Gaussian processes as an interpolation technique for geospatial trajectories. A Gaussian process models measurements of a trajectory as coming from a multidimensional Gaussian, and it produces for each timestamp a Gaussian distribution as a prediction. We discuss elements that need to be considered when applying Gaussian process to trajectories, common choices for those elements, and provide a concrete example of implementing a Gaussian process. 
\end{abstract}


\section{Introduction}
The availability of devices with location tracking capability has helped generate a tremendous amount of trajectory data of humans, animals, vehicles, and drones, which is valuable in various applications~\cite{zheng2015trajectory}. However, trajectories may have missing locations between their measurements and uncertainty in the measurements depending on how they were captured. 
Interpolating missing locations between measurements, or making predictions about future locations, is often necessary in order to derive high utility from trajectories. This work presents Gaussian Process (GP) as a powerful and flexible technique for trajectory interpolation and prediction. 

\begin{figure}[htbp!]
\centering
\includegraphics[width=.5\linewidth]{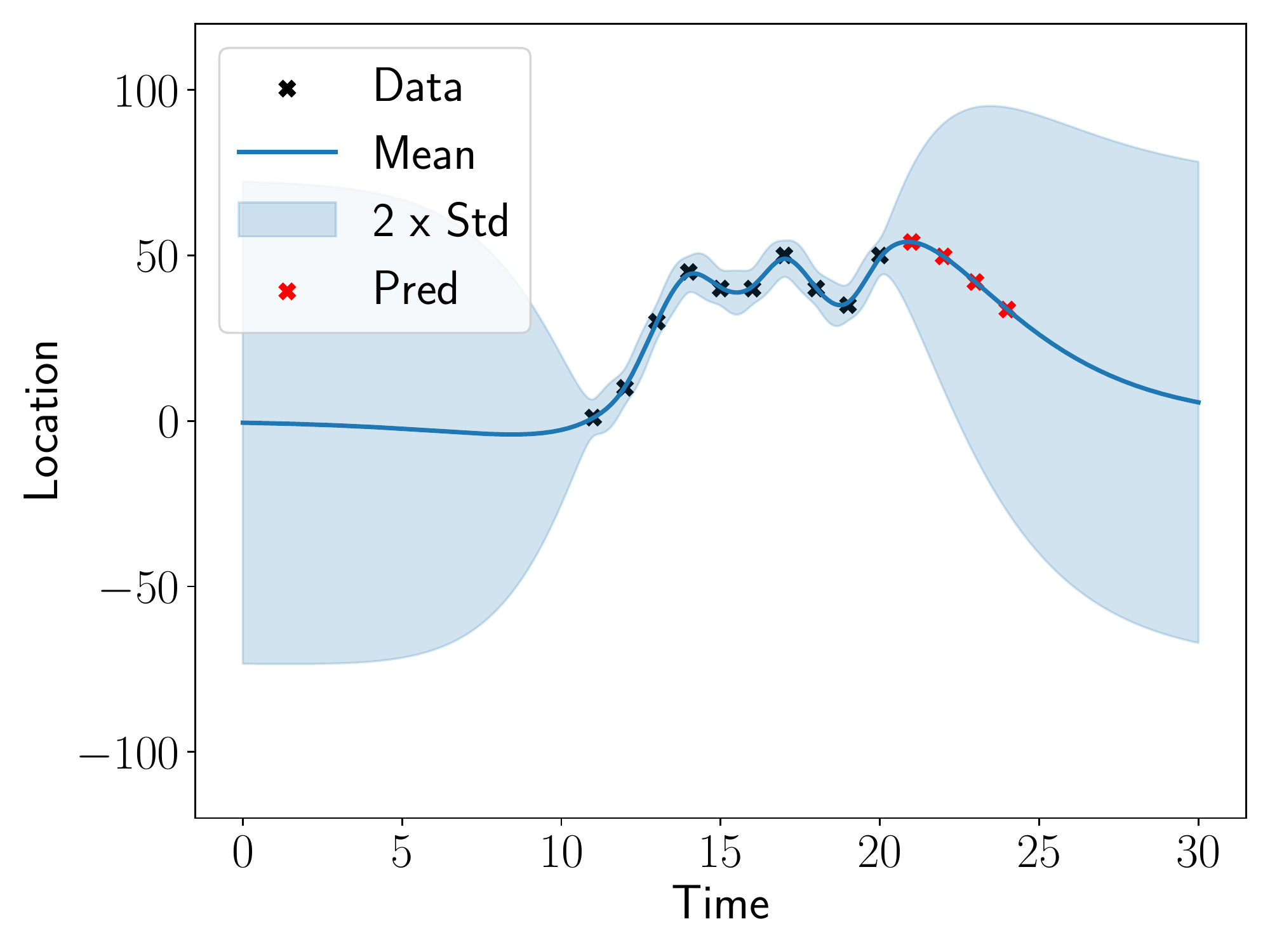}
\caption{Illustration of a Gaussian process trained from data (black cross) making predictions (red cross).}  
\label{fig:GP_illustration}
\end{figure}

A trajectory $\traj$ is defined as a sequence of time-ordered noisy location measurements $\traj = \{ \dpoint_1, \dpoint_2, \dots, \dpoint_{|\traj|} \}$, where each measurement or data point $\dpoint$ (bold symbol) includes $\dpoint.\dpointlon$ and $\dpoint.\dpointlat$ as the longitude and latitude, $\dpoint.\dpointtime$ as the timestamp, and $\dpoint.\dpointsigma$ as the accuracy or uncertainty. Previous work~\cite{diggelen2007system} showed that it is reasonable to assume Gaussian noise for location measurements such as GPS points. Therefore,  $\dpoint.\dpointsigma$ can be considered as the standard deviation of independent Gaussian noise of longitude and latitude. Figure~\ref{fig:GP_illustration} shows an example of a trajectory with its measurements shown as black crosses, uncertainty as the blue area, and some predictions made by a GP, which is explained later.

GP has been studied and applied extensively~\cite{williams2006gaussian}. There have been several works that applied GP for different types of trajectories such as in robotics~\cite{cox2012gaussian,barfoot2014batch,hewing2020simulation,cao2017gaussian,heravi2011long}, video surveillance~\cite{kim2011gaussian,ellis2009modelling}, batch trajectories processing~\cite{tiger2015online}, sensor trajectories in active sensing~\cite{le2009trajectory}, travel-time prediction~\cite{ide2009travel}, and human motion~\cite{hong2019gaussian,wang2007gaussian}. This work describes how to apply GP for trajectory interpolation and prediction, especially GPS-based trajectories. 
\section{Gaussian Process} \label{sec:gaussian-process-general}

A Gaussian Process (GP) is a generalization of the Gaussian probability distribution. While a probability distribution describes scalar or vector random variables, a stochastic process describes properties of functions, i.e., instead of describing the probability of generating a point as a Gaussian distribution, a GP describes the probability of generating a function. If one can loosely consider a function as a very long vector, where each entry in the vector indicates the function value $\gpfunc(\dpointtime)$ at a specific input $\dpointtime$ (which is a timestamp in our case), then a GP implies that any subset of entries in that long vector is distributed according to a multidimensional Gaussian. We build a separate GP for latitude and another one for longitude, assuming that their values change independently.

As an example, one wants to estimate locations of a user at a set of timestamps $\gppredset = \{\dpointtime_1, \dpointtime_2, \dots, \dpointtime_{|\gppredset|}\}$. Thus, the input of the function $\gpfunc(\dpointtime)$ that a GP (for either latitude or longitude) describes is a timestamp $\dpointtime \in \gppredset$, and the output is the mean and variance of the Gaussian distribution that describes the distribution of $\gpfunc(\dpointtime)$. A GP implies that the set $\{\gpfunc(\dpointtime) | \dpointtime \in \gppredset \}$ is distributed according to a $|\gppredset|$-dimensional Gaussian. When $\gppredset$ includes only one timestamp $\dpointtime$, the longitude (or latitude) of the estimated location is distributed according to a univariate Gaussian distribution. Figure~\ref{fig:GP_illustration} shows an example of a GP trained from data points (represented as black crosses) and making predictions (represented as red crosses). The location axis can either be longitude, latitude, or their transformed values such as converted to another spatial coordinate system. 

A GP is specified by its mean function $\meanfunc(\dpointtime)$ and covariance function $\kernelfunc(\dpointtime, \dpointtime')$. A deterministic \textit{mean function} $\meanfunc(\dpointtime)$ is used to specify the mean of the multidimensional Gaussian, defining the mean of the estimated locations. In the literature, GPs are often studied with a zero mean function, which means $\meanfunc(\dpointtime) = 0$ for all $\dpointtime$, because a deterministic mean function can be incorporated after the kernel function is trained from training data. 

The covariance matrix of the multidimensional Gaussian defines how points at different values of the independent variable (time in our case) correlate with each other. In general, timestamps that are closer would correlate more and those that are further apart would correlate less, which means the correlation decreases to zero as $|\dpointtime-\dpointtime'|$ becomes larger. A GP models such correlation using a scalar \textit{covariance kernel/function} $\kernelfunc(\dpointtime, \dpointtime')$. A GP can be very flexible by using and combining different kernels, which will be discussed later. A kernel often has some parameters that need to be trained from data, such as pertaining to the uncertainty of the input data or how strong a correlation should be given the time difference $|\dpointtime-\dpointtime'|$.

When using a GP for trajectory interpolation and prediction, one would need to specify its mean function $\meanfunc(\dpointtime)$ and covariance function $\kernelfunc(\dpointtime, \dpointtime')$, train its parameters from data, then make predictions. We discuss each of these elements in the next section. 
\section{Gaussian Process Elements} \label{sec:elements}
In this section, we describe the elements needed to implement a GP for trajectories, including data preparation, mean function, covariance function (or kernel), training, and inference. Our implementation uses the GPFlow library~\cite{GPflow2020multioutput}, which is a package for building Gaussian process models in Python, but the concepts also apply to other libraries.

\subsection{Data Preparation}
We first describe how to prepare data from a trajectory $\traj = \{ \dpoint_1, \dpoint_2, \dots, \dpoint_{|\traj|} \}$ for GP training and inference. As mentioned, two separate GPs are created for longitude and latitude predictions. Input data for training each GP includes the sequence $\{(\dpointtime_1, \locvalue_1), \dots, (\dpointtime_{|\traj|}, \locvalue_{|\traj|})\}$ where $\dpointtime_i = \dpoint_i.\dpointtime$ and coordinate $\locvalue_i$ can be either $\dpoint_i.\dpointlon$ or $\dpoint_i.\dpointlat$, and the measurement uncertainty $\sigmam = \dpoint.\dpointsigma$. From the training sequence, we create a feature vector $\fvec = [\dpointtime_1, \dots, \dpointtime_{|\traj|}]^T$ and label vector $\lvec = [\locvalue_1, \dots, \locvalue_{|\traj|}]^T$. 
Similarly, for interpolating or predicting locations of the user at timestamps in a timestamp set $\gppredset$, a feature vector $\fvec_* = [{\dpointtime | \dpointtime \in \gppredset}]^T$ is created. 

In addition to $\fvec, \lvec, \fvec_*$ and $\sigmam$, one may also use a prior uncertainty $\sigmaprior$, which represents the prior information of the uncertainty of locations of the users in terms of the standard deviation of a distribution. For example, $\sigmaprior$ can be the standard deviation of a large Gaussian distribution covering an entire city area, which would indicate a high degree of prior uncertainty.

\subsection{Mean Function}
As mentioned, the mean function $\meanfunc(\dpointtime)$ is a deterministic function defining the mean of the estimated locations. A GP makes predictions based mainly on parameters learned from training data. When a prediction timestamp is far from training timestamps, the prediction tends to return to the mean function. 

A GP is often assumed to have zero mean, i.e., $\meanfunc(\dpointtime) = 0, \forall \dpointtime$. However, when there is a clear pattern, it can be beneficial to provide a more realistic mean to the model. For example, a trajectory may have a linear trend or we may believe future locations are likely to follow a linear extrapolation of two most recent measurements. 

Using the GPFlow library, common mean functions can be specified using \textit{gpflow.mean\_functions}. For example, a constant mean function $\meanfunc(\dpointtime) = c$ with \textit{gpflow.mean\_functions.Constant}, a linear mean function $\meanfunc(\dpointtime) = At + b$ with \textit{gpflow.mean\_functions.Linear}, or a custom mean function can also be specified. 

\subsection{Kernel/Covariance Function}
\begin{figure}[htbp!]
\centering
\includegraphics[width=0.95\linewidth]{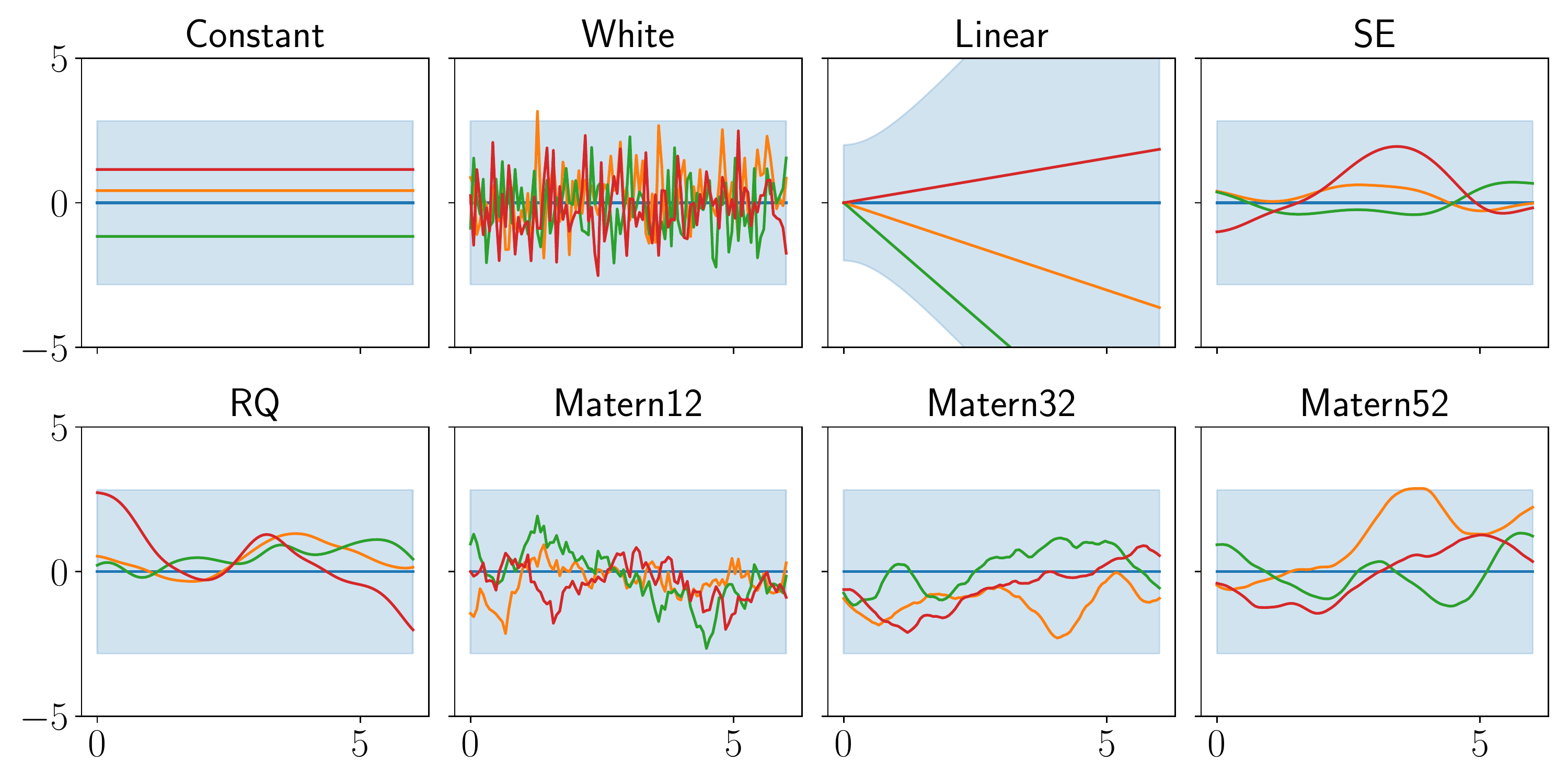}
\caption{Examples of functions drawn from GPs with different kernels. Each figure shows 3 functions drawing from the same GP with the blue line indicating the mean of the GP and the blue area indicating the variance of the GP. Predictions made from these kernels would have a similar shape.}  
\label{fig:kernels_illustration}
\end{figure}

The kernel or covariance function $\kernelfunc(\dpointtime, \dpointtime')$ is the core element that helps a GP capture trends in data and gives a GP flexibility and efficiency. Different kernels capture different kinds of trends, and kernels can also be combined together for more complex models. In this section, we describe some common kernels and some common types of combinations. Examples of these kernels are shown in Figure~\ref{fig:kernels_illustration}, and examples of these combinations are shown in Figure~\ref{fig:kernel_combination_illustration}. More detail about kernels can be found in~\cite{williams2006gaussian} and~\cite{duvenaud2021gpkernelcookbook}.

Note that parameters of a kernel can be set to be trained from data or fixed. For example, if we know the measurement uncertainty $\sigmam$, we can set it to be fixed; otherwise, it can be learned from the data. If using the GPFlow library, this can be set using the \textit{set\_trainable} method. 

\subsubsection{Common Kernel Types}

\paragraph{Constant Kernel}
The constant kernel outputs constant predictions, i.e., $\gpfunc(\dpointtime) = c$ with $c \sim \normal(0, \sigma^2)$. The formula is 
\begin{equation}
    \kernelfunc_{Constant}(\dpointtime, \dpointtime') = \sigma^2
\end{equation}
The constant kernel is often used to modify the mean of a GP or to scale the magnitude of the other factor (kernel) of a kernel. The GPFlow library provides this kernel by \textit{gpflow.kernels.Constant} with parameter \textit{variance}. 

\paragraph{White Kernel}
The white kernel explains/produces the noise-component of a measurement. The formula is 
\begin{equation}
    \kernelfunc_{White}(\dpointtime, \dpointtime') = \sigma^2 \delta(\dpointtime, \dpointtime')
\end{equation}
where $\delta(\dpointtime, \dpointtime') = 1$ when $\dpointtime = \dpointtime'$; otherwise, $\delta(\dpointtime, \dpointtime') = 0$. GPFlow provides this kernel with \textit{gpflow.kernels.White} with parameter \textit{variance}. Since we know that the measurement uncertainty is $\sigmam$, we can set the variance value of this kernel to $\sigmam^2$ (measurement noise) and set this parameter as non-trainable.

\paragraph{Linear Kernel}
The linear kernel outputs linear predictions, i.e., $\gpfunc(\dpointtime) = c\dpointtime$ with $c \sim \normal(0, \sigma^2)$. The formula is 
\begin{equation}
    \kernelfunc_{Linear}(\dpointtime, \dpointtime') = \sigma^2 \dpointtime \dpointtime'
\end{equation}
The linear kernel is often used in combination with other kernels to capture linear trends in data. This kernel is a \textit{non-stationary} kernel, which means it depends on the absolute locations of the two inputs. We will see later some \textit{stationary} kernels, which only depend on the relative positions of its two inputs. The GPFlow library provides this kernel by \textit{gpflow.kernels.Linear} with parameter \textit{variance}.

\paragraph{Squared Exponential (SE)/Radial Basis Function (RBF) Kernel}
The SE/RBF kernel is often used as the default kernel for a GP. This is a stationary and smooth kernel with the formula
\begin{equation}
    \kernelfunc_{SE}(\dpointtime, \dpointtime') = \sigma^2 \exp{\Big(-\frac{(\dpointtime - \dpointtime')^2}{2l^2}\Big)}
\end{equation}
The length scale $l$ indicates how much one data point affects another and is often trained from data. The variance $\sigma^2$ is trained from data but one can set this variance to $\sigmaprior^2$ to provide prior information about the variance of predictions.
GPFlow provides this kernel by \textit{gpflow.kernels.SquaredExponential} with parameters \textit{variance} and \textit{length\_scale}. Also, instead of providing scalar values, prior distributions can also be used as the prior for the length scale $l$ and  variance $\sigma^2$.

\paragraph{Rational Quadratic (RQ) Kernel}
Another common kernel that is often used as the default kernel for a GP is the rational quadratic (RQ) kernel with the formula
\begin{equation}
    \kernelfunc_{RQ}(\dpointtime, \dpointtime') = \sigma^2 \Big(1 + \frac{(\dpointtime - \dpointtime')^2}{2 \alpha l^2}\Big)^{-\alpha}
\end{equation}
The RQ kernel is equivalent to adding together many SE kernels with different length scales $l$. It converges to the SE kernel when $\alpha \rightarrow \infty$. Parameter $\alpha$ controls the relative weighting of large-scale and small-scale variations. The GPFlow library provides this kernel by \textit{gpflow.kernels.RationalQuadratic} with parameters \textit{variance}, \textit{length\_scale}, and \textit{alpha}.

\paragraph{Matérn Kernel}
The Matérn kernel is another common kernel. It is a stationary kernel and a generalization of the SQ kernel with an additional parameter $\nu$ controlling the smoothness. It converges to the SE kernel when $\nu \rightarrow \infty$. The formula for the Matérn kernel is rather complicated. However, there are three common values of $\nu$, which are $\frac{1}{2}, \frac{3}{2}, \frac{5}{2}$, that produce three common kernels called \textit{Matern12}, \textit{Matern32}, and \textit{Matern52}. 
Their formulas are
\begin{align}
    \kernelfunc_{Matern12}(\dpointtime, \dpointtime') &= \sigma^2 \exp{\Big(-\frac{1}{l} |\dpointtime - \dpointtime'| \Big)}\\
    \kernelfunc_{Matern32}(\dpointtime, \dpointtime') &= \sigma^2 \Big(1+\frac{\sqrt{3}}{l} |\dpointtime - \dpointtime'| \Big) \exp{\Big(-\frac{\sqrt{3}}{l} |\dpointtime - \dpointtime'| \Big)}\\
    \kernelfunc_{Matern52}(\dpointtime, \dpointtime') &= \sigma^2 \Big(1+\frac{\sqrt{5}}{l} |\dpointtime - \dpointtime'| + \frac{5}{3l} (\dpointtime - \dpointtime')^2 \Big) \exp{\Big(-\frac{\sqrt{5}}{l} |\dpointtime - \dpointtime'| \Big)}
\end{align}
Using the GPFlow library, these kernels are provided by \textit{gpflow.kernels.Matern12}, \textit{gpflow.kernels.Matern32} and \textit{gpflow.kernels.Matern52} with parameters \textit{variance} and \textit{length\_scale}.

\subsubsection{Common Types of Kernel Combination}
\begin{figure}[htbp!]
\centering
\includegraphics[width=0.95\linewidth]{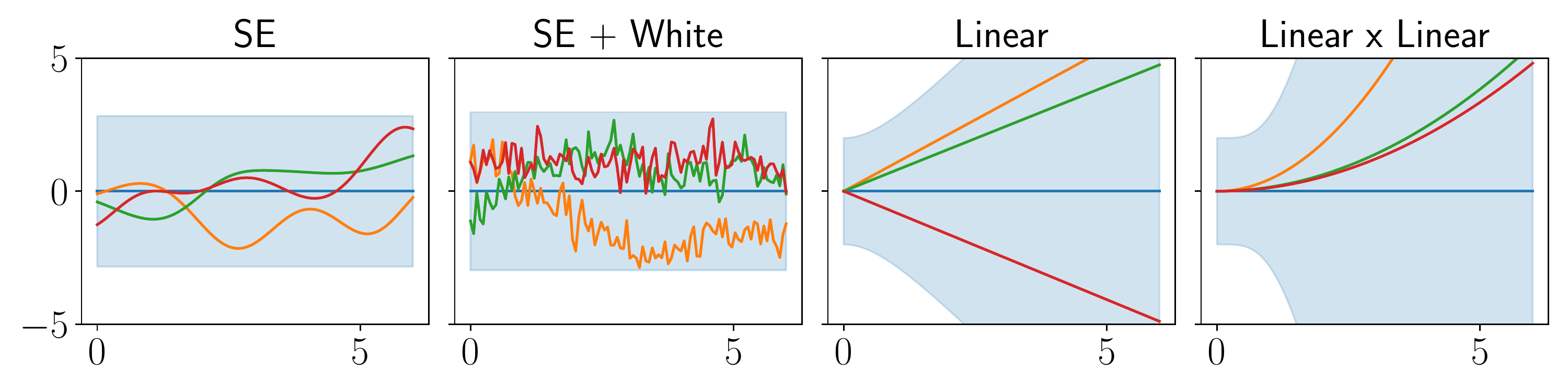}
\caption{Examples of functions drawn from GPs with different types of kernel combination. Each figure shows 3 functions drawing from the same GP with the blue line indicating the mean of the GP and the blue area indicating the variance of the GP. Predictions made from these kernels would have a similar shape.}  
\label{fig:kernel_combination_illustration}
\end{figure}

Kernels can be combined together to create more complex kernels, helping to capture more complex trends in the data. For example, we may want to capture both a wiggling movement with an SE kernel and measurement uncertainty with a white kernel. Two common types of combining kernels are summing and multiplying. Examples of these combinations are shown in Figure~\ref{fig:kernel_combination_illustration}.

\paragraph{Summing Kernels}
Summing two kernels $\kernelfunc_1$ and $\kernelfunc_2$ means summing their output:
\begin{equation}
    \kernelfunc_{Sum}(\dpointtime, \dpointtime') = \kernelfunc_{1}(\dpointtime, \dpointtime') + \kernelfunc_{2}(\dpointtime, \dpointtime')
\end{equation}
Roughly speaking, summing kernels is similar to an OR operation, which means the output of $\kernelfunc_{Sum}$ would be higher if the output of either $\kernelfunc_{1}$ or $\kernelfunc_{2}$ is higher. An example of summing kernels is to capture both a wiggling movement with a SE kernel and measurement uncertainty with a white kernel:
\begin{equation}
    \kernelfunc(\dpointtime, \dpointtime') = \kernelfunc_{SE}(\dpointtime, \dpointtime') + \kernelfunc_{White}(\dpointtime, \dpointtime') = \sigma_{SE}^2 \exp{\Big(-\frac{(\dpointtime - \dpointtime')^2}{2l^2}\Big)} + \sigma_{White}^2 \delta(\dpointtime, \dpointtime')
\end{equation}
An illustration of this combination is shown in Figure~\ref{fig:kernel_combination_illustration}.

\paragraph{Multiplying Kernels}
Multiplying two kernels $\kernelfunc_1$ and $\kernelfunc_2$ means multiplying their output:
\begin{equation}
    \kernelfunc_{Mul}(\dpointtime, \dpointtime') = \kernelfunc_{1}(\dpointtime, \dpointtime') \times \kernelfunc_{2}(\dpointtime, \dpointtime')
\end{equation}
Roughly speaking, multiplying kernels is similar to an AND operation, which means the output of $\kernelfunc_{Sum}$ would be higher if the output of both $\kernelfunc_{1}$ and $\kernelfunc_{2}$ is higher. An example of multiplying kernels is multiplying two linear kernels to create a quadratic kernel or more to create higher-degree kernels. An illustration of this combination is shown in Figure~\ref{fig:kernel_combination_illustration}.

\subsection{Training and Inference}
After constructing a mean function and a kernel and setting trainable parameters appropriately, training and inference processes are often straightforward using provided methods in GP libraries. Using GPFlow, for training, one can use existing optimizers in the package \textit{gpflow.optimizers} such as the \textit{Scipy} optimizer with input $\fvec, \lvec$. 
Once trained, we can use the trained model to make predictions with \textit{predict\_y}, which returns the mean and variance of predictions for new data points $\fvec_*$. 

Note that the training process minimizes the negative log marginal likelihood. Since likelihood specifies the probability density of the observations/measurements given the parameters, for trajectory data, we can use Gaussian likelihood, which is the default in the GPFlow library. The likelihood variance indicates the uncertainty learned from data. For trajectory data, uncertainty in the movement of measurements can be captured by the model with likelihood variance, and the uncertainty of each measurement can be captured by a white kernel.
\section{Gaussian Process Example} \label{sec:example}
In this section, we present an example of using GP for trajectories.
In this example, we use given (training) timestamps $\fvec = [11,  12, \dots, 20]^T$, given locations $\lvec = [1, 10, 30, 45, 40, 40, 50, 40, 35, 50]^T$, and seek computed values at timestamps $\fvec_* = [21,  22, \dots, 24]^T$ (illustrated in Figure~\ref{fig:GP_illustration}). For reproducibility, we use the GPFlow library and set the TensorFlow random seed to 1. GPFlow also provides a utility method \textit{print\_summary} that prints values of the parameters of a model.

\begin{figure*}[htbp!]
\centering
\begin{subfigure}{0.3\linewidth}
  \centering\captionsetup{justification=centering}
  \includegraphics[width=\textwidth]{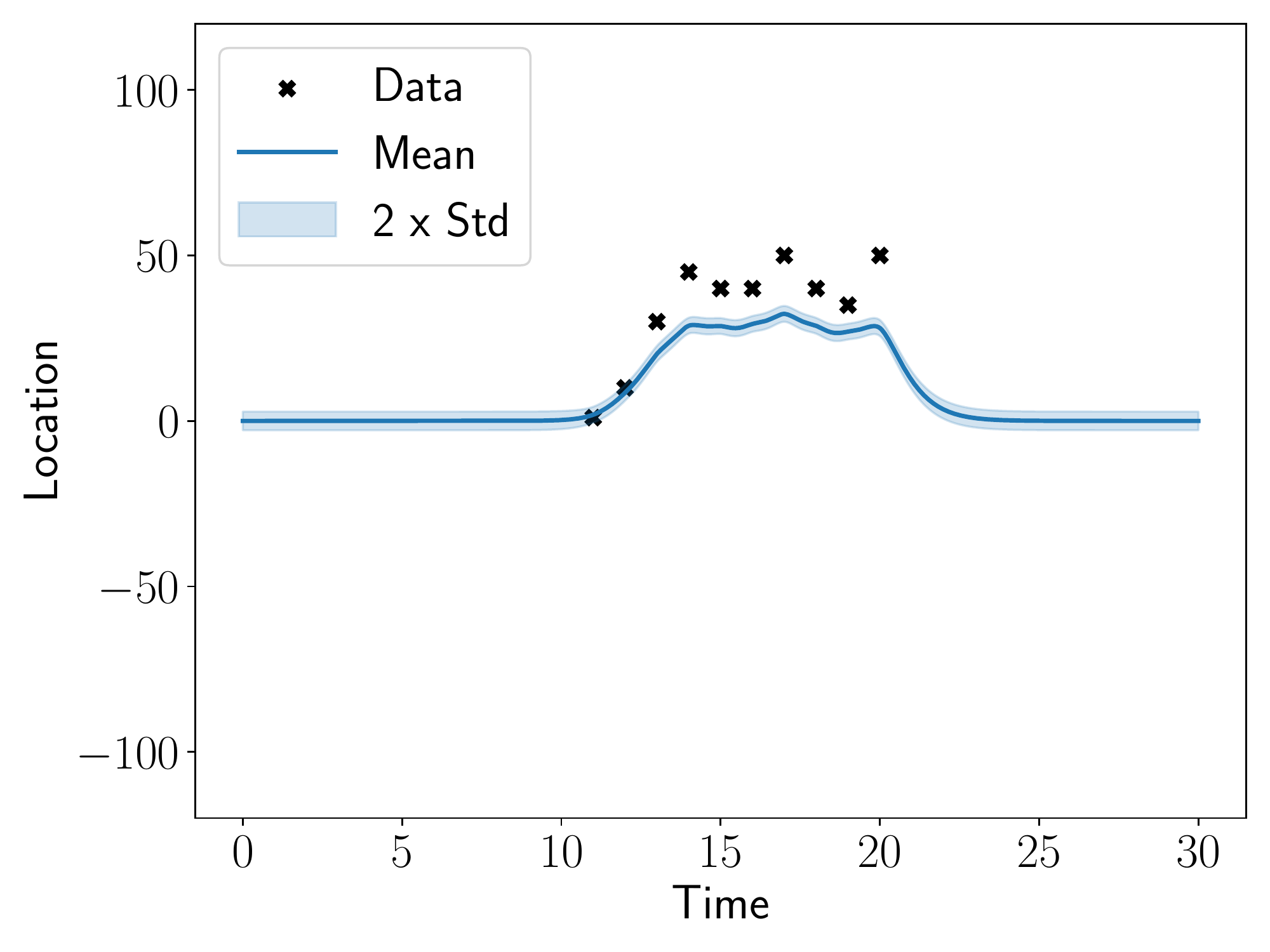}
  \caption{Not trained}
  \label{fig:GP_examples_not_trained}
\end{subfigure}%
\begin{subfigure}{.3\linewidth}
  \centering\captionsetup{justification=centering}
  \includegraphics[width=\textwidth]{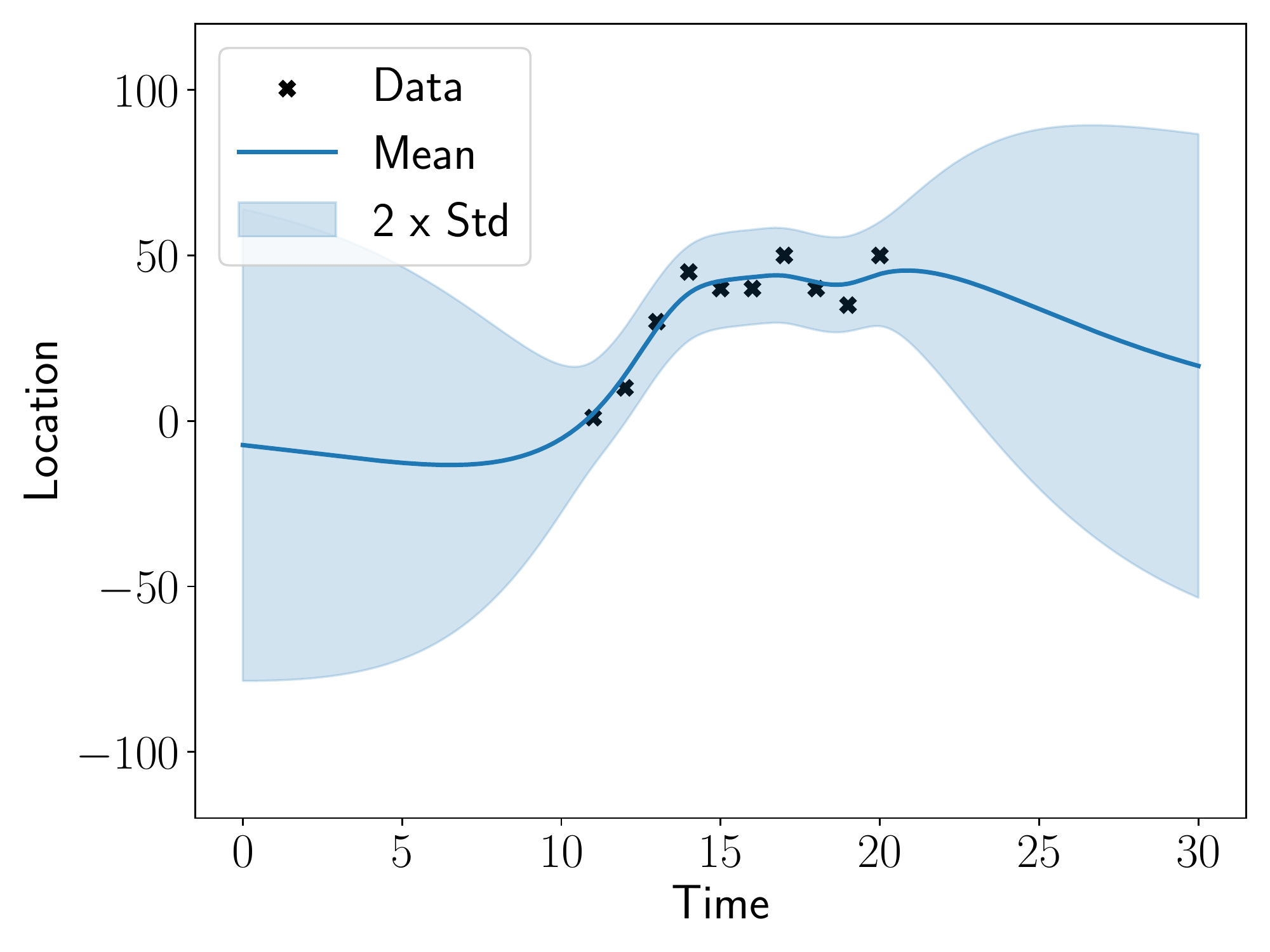}
  \caption{Trained}
  \label{fig:GP_examples_trained}
\end{subfigure}%
\begin{subfigure}{.3\linewidth}
  \centering\captionsetup{justification=centering}
  \includegraphics[width=\textwidth]{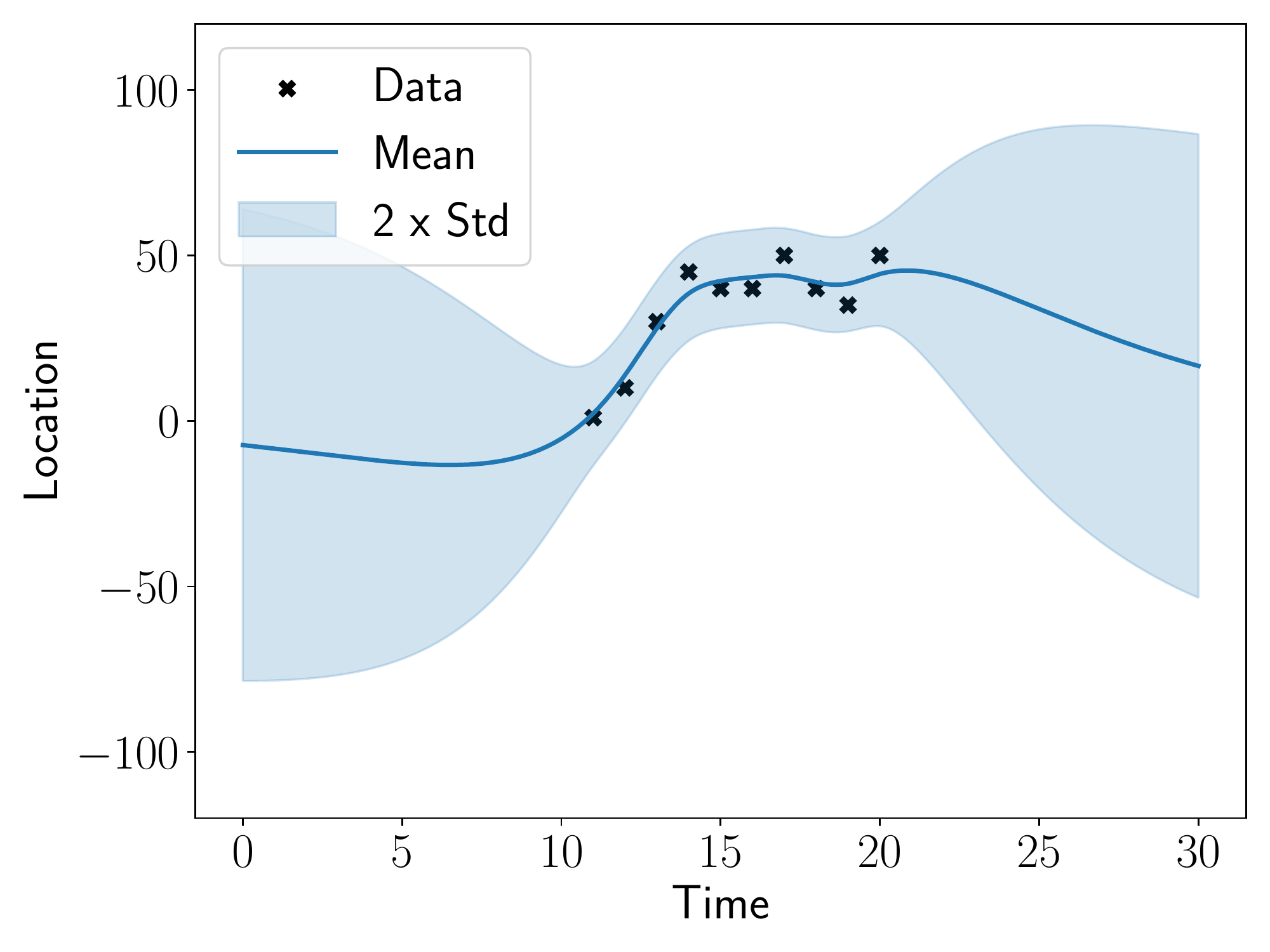}
  \caption{Trained with White noise}
  \label{fig:GP_examples_trained_w_white}
\end{subfigure}
\begin{subfigure}{.3\linewidth}
  \centering\captionsetup{justification=centering}
  \includegraphics[width=\textwidth]{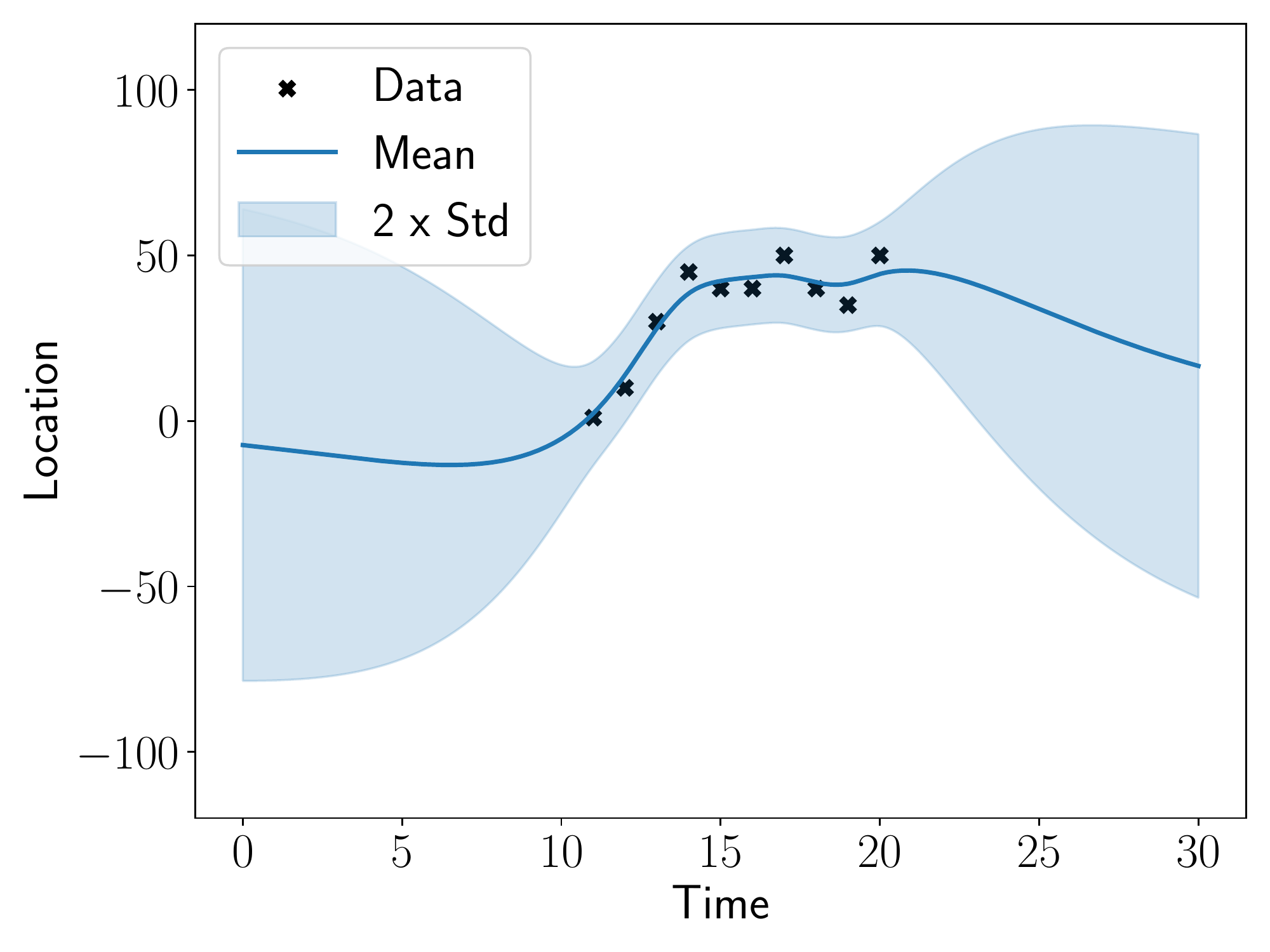}
  \caption{Same as (\protect\subref{fig:GP_examples_trained_w_white}), but with fixed White noise}
  \label{fig:GP_examples_trained_w_white_nontrainable}
\end{subfigure}%
\begin{subfigure}{.3\linewidth}
  \centering\captionsetup{justification=centering}
  \includegraphics[width=\textwidth]{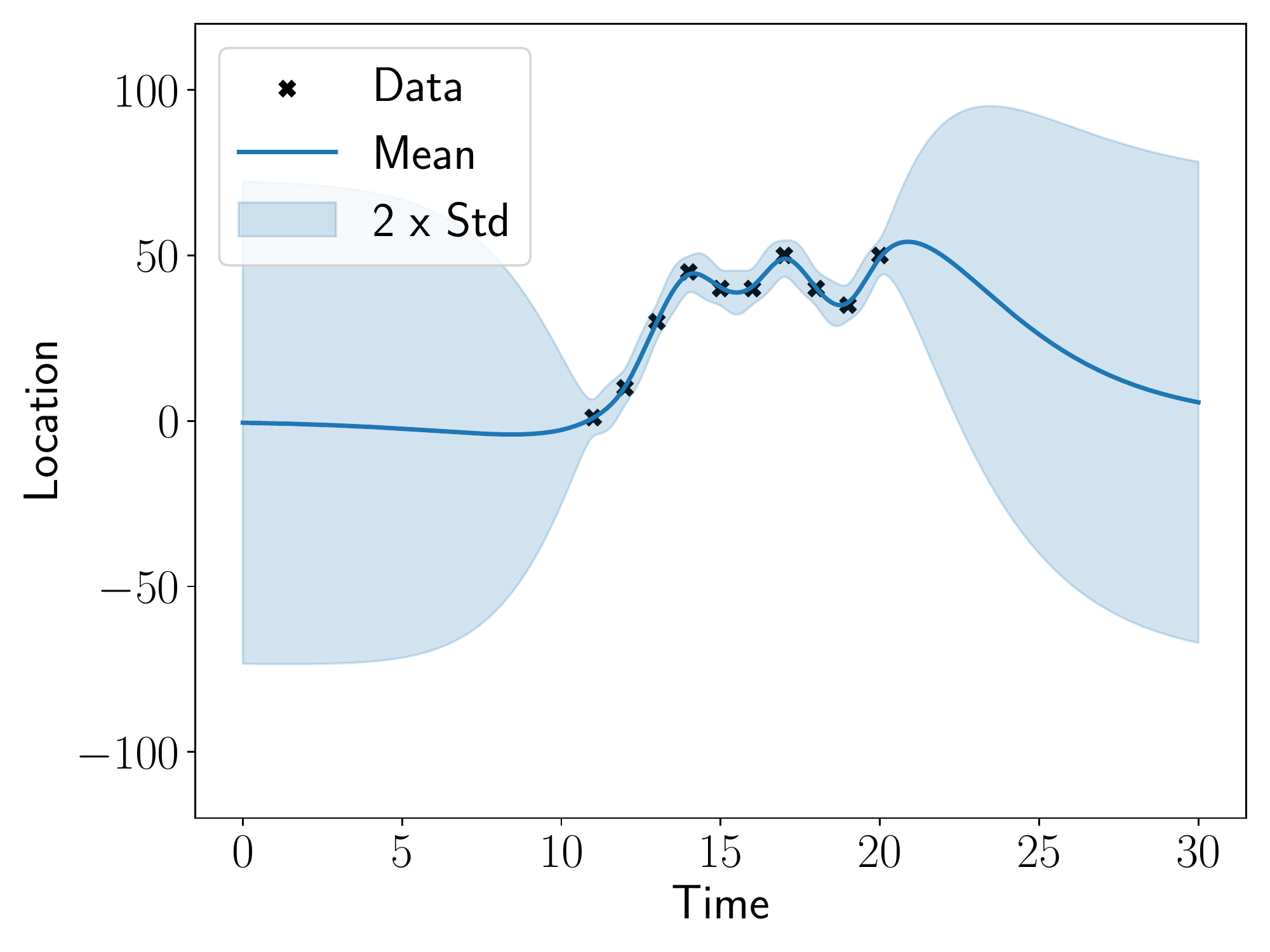}
  \caption{Same as (\protect\subref{fig:GP_examples_trained_w_white_nontrainable}), but with small likelihood variance}
  \label{fig:GP_examples_trained_w_white_nontrainable_small_likelihood_var}
\end{subfigure}%
\begin{subfigure}{.3\linewidth}
  \centering\captionsetup{justification=centering}
  \includegraphics[width=\textwidth]{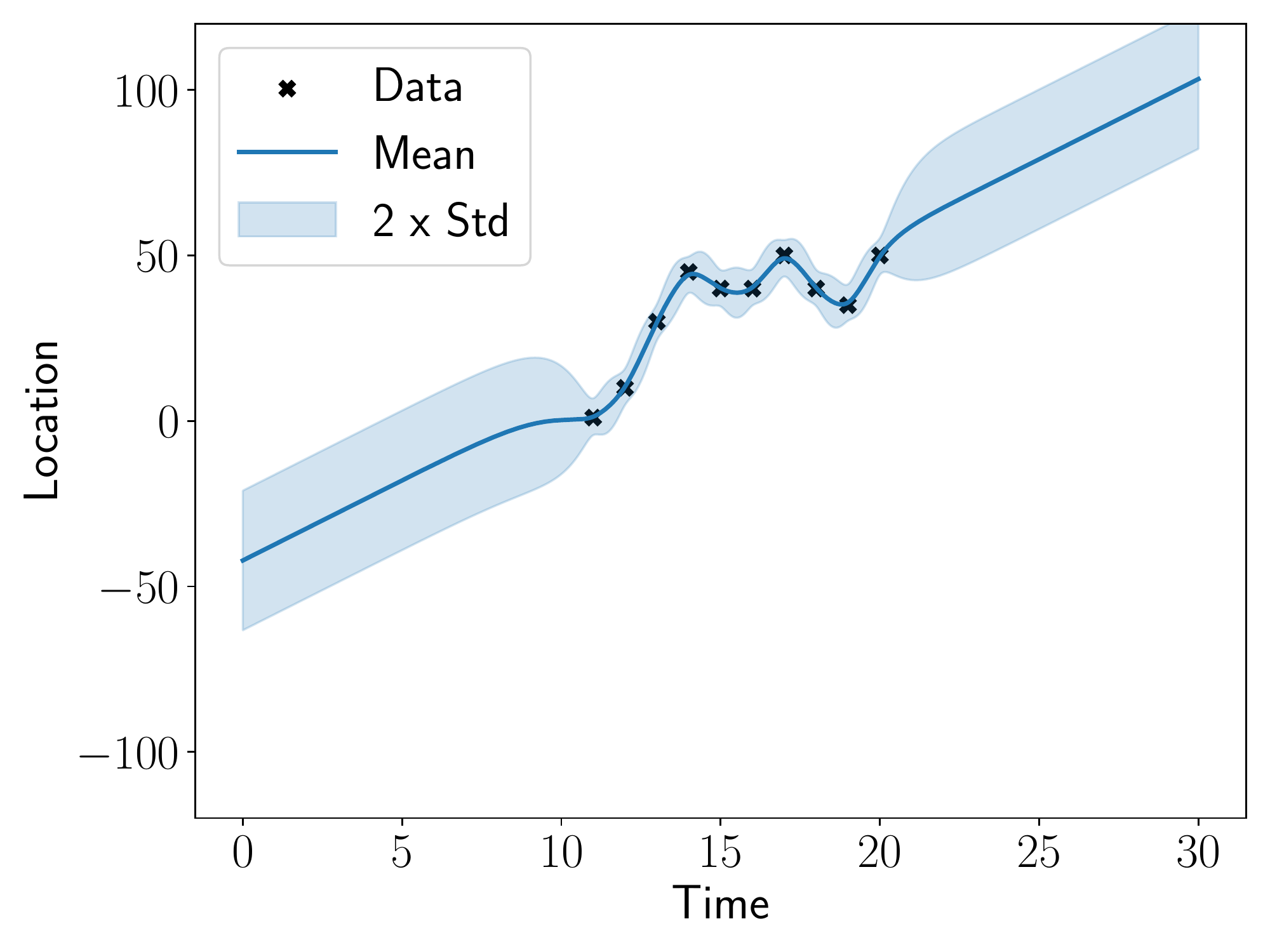}
  \caption{Same as (\protect\subref{fig:GP_examples_trained_w_white_nontrainable_small_likelihood_var}), but with linear mean function}
  \label{fig:GP_examples_trained_w_white_nontrainable_small_likelihood_var_linear_mean_f}
\end{subfigure}
\caption{Example of GPs trained from the same training data (as black cross) with different configuration of kernel and mean function.}
\label{fig:GP_examples}
\end{figure*}

The first model $\model_0$ has a zero mean function and a Matern32 kernel with default parameters, which is \textit{lengthscales} $l = 1$ and \textit{variance} $\sigma^2 = 1$. Figure~\ref{fig:GP_examples_not_trained} shows the predictions of this model before we trained its parameters with $\fvec$. We can see that the predictions do not appropriately capture measurements and give a quick return to 0. 

After we train the model with the data to get a trained model $\model_1$, we have $l = 7.35$, $\sigma^2 = 1346.36$, and likelihood variance (which indicates model variance) $\sigma_{likelihood}^2 = 36.4$. Figure~\ref{fig:GP_examples_trained} shows the predictions of this model after we trained with $\fvec$. It is clear that $\model_1$ can capture the data and trend in the data significantly better than $\model_0$.

Next, we consider model $\model_2$ as a summed combination of a Matern32 kernel and a white kernel, all with default parameters indicating default priors (e.g., prior for measurement uncertainty captured by the white kernel is 1). This also means all parameters can be trained from data.
After we train the model with the data, we have $l = 7.35$, $\sigma_{Matern32}^2 = 1346.36$, $\sigma_{White}^2 = 18.23$, and $\sigma_{likelihood}^2 = 18.23$.  Figure~\ref{fig:GP_examples_trained_w_white} shows that the model $\model_2$ makes similar predictions compared to model $\model_1$. However, the white kernel captured some of the measurement noise.

Next, we consider model $\model_3$, which similar to $\model_2$, but we fix the variance of the white kernel to \textit{variance} $\sigma_{White}^2 = 4$, which means we assume that each measurement has uncertainty $\dpointsigma^2 = 4$. Fixing the variance means the training process will not change this parameter.
After we train the model with the data, we have $l = 7.35$, $\sigma_{Matern32}^2 = 1346.36$, $\sigma_{White}^2 = 4$, and $\sigma_{likelihood}^2 = 32.464$. Since we fixed the variance of the white kernel, the likelihood variance was adjusted to capture more of the uncertainty.
Figure~\ref{fig:GP_examples_trained_w_white_nontrainable} shows that the model $\model_3$ make similar predictions compared to model $\model_2$.

Next, we consider model $\model_4$ similar to $\model_3$ but in addition to fixing the variance of the white kernel to $4$, we also set the likelihood variance to a very small initial value $\sigma_{likelihood}^2 = 0.0001$, which indicates that we are confident of our knowledge of measurement uncertainty.
After we train the model with the data, we have $l = 4.11$, $\sigma_{Matern32}^2 = 1328.03$, $\sigma_{White}^2 = 4$, and $\sigma_{likelihood}^2 = 0.00015$. 
Figure~\ref{fig:GP_examples_trained_w_white_nontrainable_small_likelihood_var} shows that, compared to model $\model_3$, the model $\model_4$ make predictions closer to the training measurements. Figure~\ref{fig:GP_illustration} shows predictions made by $\model_4$ for $\fvec_*$.

In the final example, we consider model $\model_5$ similar to $\model_4$ but using a linear mean function starting with a guess $y = x + 1$.
After we train the model with the data, we have $l = 1.37$, $\sigma_{Matern32}^2 = 107.24$, $\sigma_{White}^2 = 4$, $\sigma_{likelihood}^2 = 0.0001$, and a mean function $y = 4.84x -42.14$. The training process learned the linear line going through the measurements and used that as the mean function. The model then made predictions closer to that mean function and with smaller variance, as shown in Figure~\ref{fig:GP_examples_trained_w_white_nontrainable_small_likelihood_var_linear_mean_f}.
\section{Discussion} \label{sec:discussion}
Gaussian process is a powerful and flexible technique that is capable of capturing trends and uncertainty of location measurements. A GP can use a simple or complex mean function, or even use output of other models, such as a deep neural network model~\cite{fortuin2019deep}, as its mean function. A GP can use standard kernels, combined kernels, or other complex kernels. These capabilities help GP incorporate knowledge from other domains for better interpolation and prediction. 

There are some issues that need to be taken into account when one considers using GPs for trajectories. One issue is that the length scale of a kernel, which roughly indicates how far in time a measurement affects other predictions, tends to be determined by the non-smooth region in the data. For trajectories, these regions are often where we have dense measurements with many direction changes. These regions can make the length scale become too small, resulting in a very sharp change in variance. It can be beneficial to create different GPs when the trajectory sampling rate changes significantly. Other issues can be the assumption of independence between longitude and latitude changes, choosing appropriate kernels, or computational complexity. However, in general, GP is a useful technique for trajectory interpolation and prediction.

\bibliographystyle{unsrt}  
\bibliography{references}  %

\end{document}